\documentclass[letterpaper, 10 pt, conference]{ieeeconf}  

\IEEEoverridecommandlockouts                              

\usepackage{graphics} 
\usepackage{epsfig} 
\usepackage{mathptmx} 
\usepackage{times} 
\usepackage{amsmath} 
\usepackage{amssymb}  
\usepackage{bbding}

\usepackage{gensymb}
\usepackage{cite}
\usepackage{algorithmic}
\usepackage{textcomp}
\usepackage{scrextend}
\usepackage{changepage}
\usepackage{multirow}
\usepackage{amsfonts}
\usepackage{booktabs}
\usepackage{mathtools}
\usepackage{xcolor}
\usepackage{float}
\usepackage{makecell}

\usepackage{hyperref}

\usepackage[linesnumbered, ruled]{algorithm2e}

\title{\LARGE \bf
Prepare the Chair for the Bear! Robot Imagination \\ of Sitting Affordance to Reorient Previously Unseen Chairs
}

\author{Xin Meng$^{1}$, Hongtao Wu$^{2}$,  Sipu Ruan$^{1}$, Gregory S. Chirikjian$^{1}$
\thanks{This work was supported by NUS Startup grants A-0009059-02-00, A-0009059-03-00, CDE Board account E-465-00-0009-01,  and National Research Foundation, Singapore, under its Medium Sized Centre Programme - Centre for Advanced Robotics Technology Innovation (CARTIN), sub award A-0009428-08-00. Hongtao Wu was supported by G. Chirikjian internal discretionary funds at Johns Hopkins University during his Ph.D. studies.}
\thanks{$^{1}$X. Meng, S. Ruan, and G. S. Chirikjian are with the Department of Mechanical Engineering, National University of Singapore, Singapore. {\tt\small\{mengxin, ruansp, mpegre\}@nus.edu.sg}.}%
\thanks{$^{2}$H. Wu is with the Laboratory for Computational Sensing and Robotics (LCSR), Johns Hopkins University, Baltimore, MD 21218, USA
        {\tt\small hwu67@jhu.edu}.}%
}

\begin{document}

\maketitle
\thispagestyle{empty}
\pagestyle{empty}

\begin{abstract}
In this letter, a paradigm for the classification and manipulation of previously unseen objects is established and demonstrated through a real example of chairs.
We present a novel robot manipulation method, guided by the understanding of object stability, perceptibility, and affordance, which allows the robot to prepare previously unseen and randomly oriented chairs for a teddy bear to sit on.
Specifically, the robot encounters an unknown object and first reconstructs a complete 3D model from perceptual data via active and autonomous manipulation.
By inserting this model into a physical simulator (\textit{i.e.}, the robot's ``imagination"), the robot assesses whether the object is a chair and determines how to reorient it properly to be used, \textit{i.e.}, how to reorient it to an upright and accessible pose.
If the object is classified as a chair, the robot reorients the object to this pose and seats the teddy bear onto the chair.
The teddy bear is a proxy for an elderly person, hospital patient, or child.
Experiment results show that our method achieves a high success rate on the real robot task of chair preparation.
Also, it outperforms several baseline methods on the task of upright pose prediction for chairs.

\end{abstract}

\section{INTRODUCTION}
\label{sec:introduction}
As robots begin to enter human life, we expect them to interact with unknown objects intelligently to help humans with daily household tasks.
This requires robots to understand: 1) \textit{what} the potential functionality an object possesses, 2) \textit{where} the object can afford such functionality, and 3) \textit{how} the functionality can be afforded.
The concept of object affordance\cite{gibson1979ecological} describes how a human might interact with an object in a particular environment so as to achieve a goal.
It encompasses the variation in functionalities\cite{bogoni1995interactive} an object can have in different scenarios.
For example, a chair possesses sitting functionality, and it can only afford such functionality when it is placed upright in an open space.
That is, when it is flipped over or blocked by obstacles, humans cannot sit on it, and thus the chair cannot afford the sitting functionality.

\begin{figure}
    \centering
    \includegraphics[width=0.9\columnwidth]{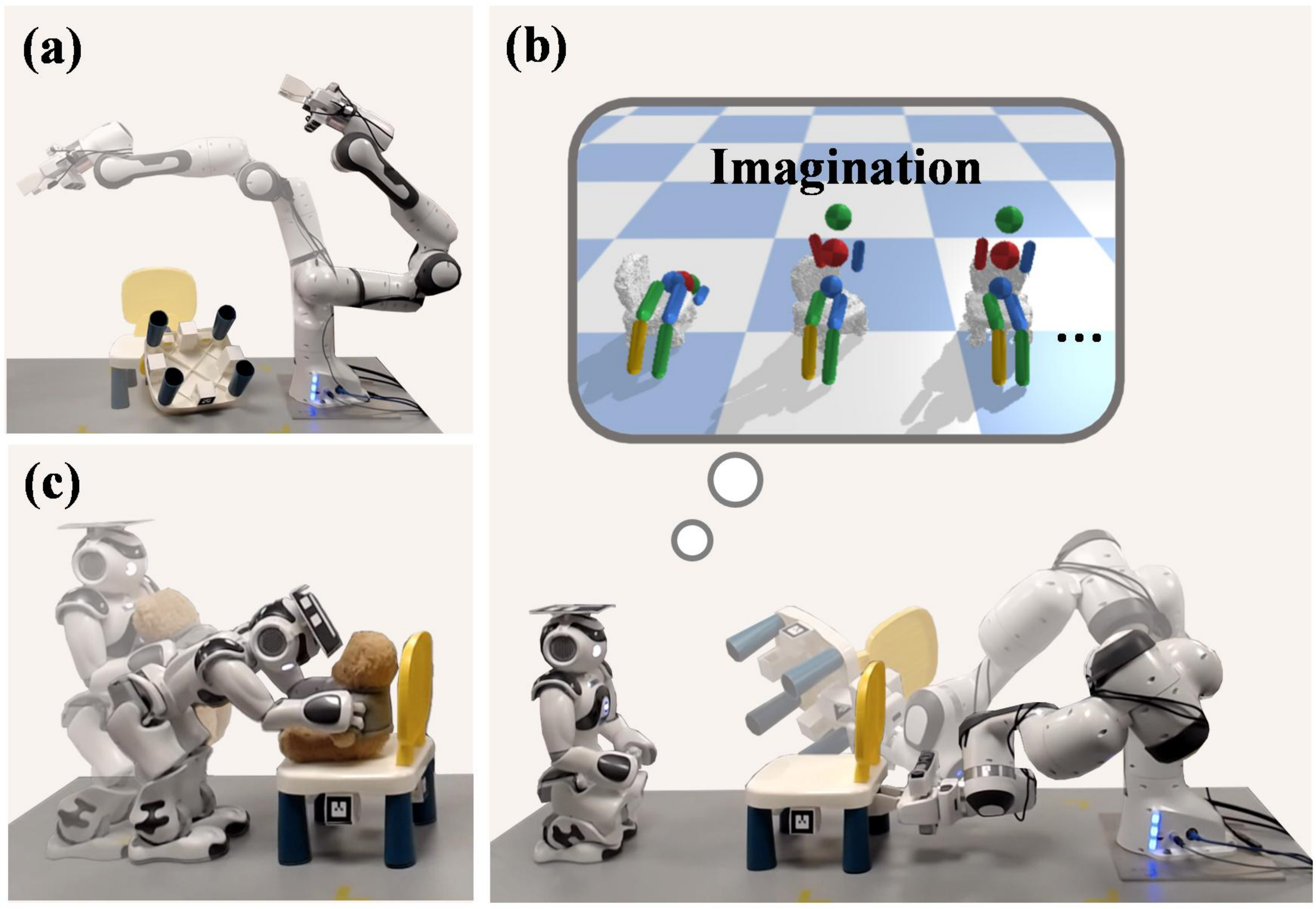}
    \caption{Overview. (a) The robot places the object into different poses to reveal the occluded part for reconstruction. (b) The robot imagines the sitting affordance of the object to determine if the object is a chair, the functional pose of the object, and how a humanoid figure can be seated. Then the robot rotates the object to the imagined pose for sitting. (c) The robot seats a teddy bear on the object according to the imagination. Video demo and more details are available at \url{https://xinnmeng.github.io/preparechair/}.}
    \label{fig: introduction}
\end{figure}

To reason about object affordances for robot-object interaction, we define objects from a robot-centric perspective via their interaction-based definition (IBD)\cite{wu2020chair}. 
For chairs, the IBD is given by: ``an object which can be stably placed on a flat horizontal surface in such a way that a typical human is able to \textit{sit}\footnote{To adopt or rest in a posture in which the body is supported on the buttocks and thighs and the torso is more or less upright. \url{https://www.collinsdictionary.com/dictionary/english/sit}} on it stably above the ground. "
This defines an object based on how it can be used.
It helps the robot to classify the object more intelligently based on \textit{what} its potential functionality is.
We define the pose that enables the object to afford the functionality as the \textit{functional pose} (\textit{i.e.}, upright pose for chairs), which answers the \textit{where} question.
When an agent sits on a chair, the body configuration and pose show \textit{how} the chair can be used for sitting.
We define the body configuration and pose as the \textit{sitting configuration} and \textit{sitting pose} of the chair.

\begin{figure*}
    \centering
    \includegraphics[width=2\columnwidth]{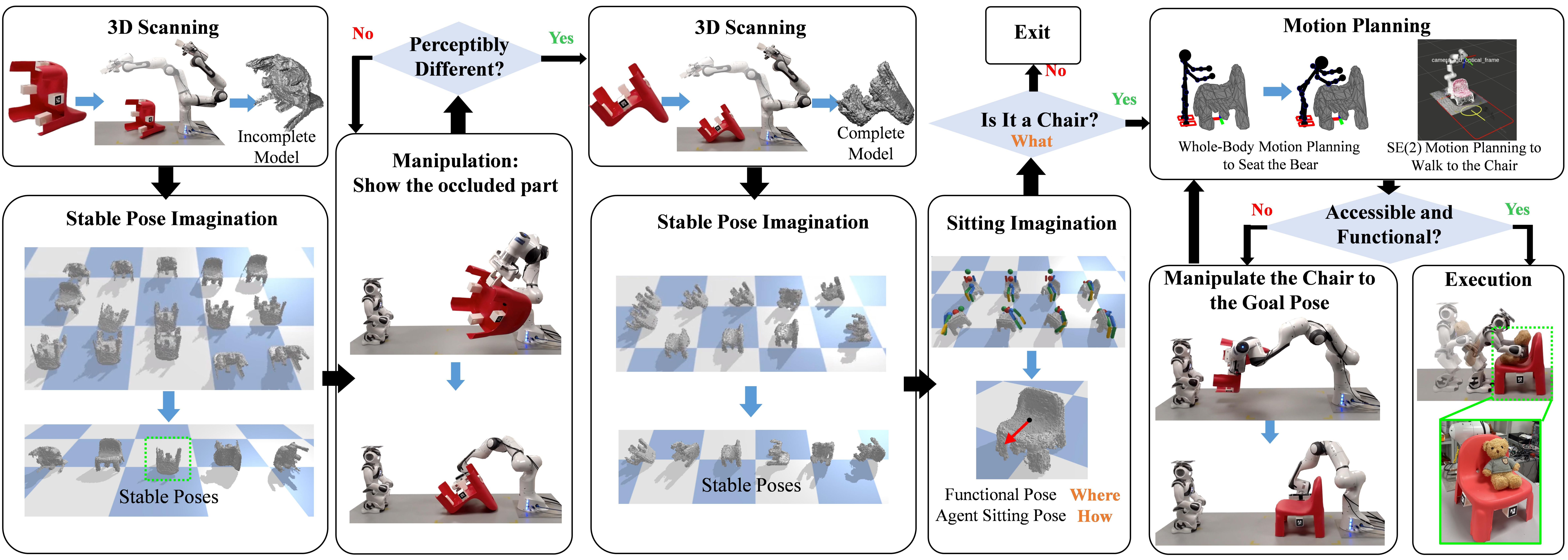}
    \caption{Pipeline. The robot arm first scans the object and performs stable pose imagination on the reconstructed model. It rotates the object to other stable poses until the object shows the occluded part. It scans the object again and generates the complete model. The robot performs stable pose imagination and sitting imagination on the complete model to determine whether the object is a chair and find a functional pose and sitting pose. The motion of the humanoid robot walking to the chair and seating the bear is planned. The robot uses the planning information to determine if the chair is accessible and/or find an accessible chair pose. The robot arm manipulates the chair to a functional and accessible pose. The humanoid then seats the bear on the chair according to the imagined sitting pose.}
    \label{fig: pipeline}
\end{figure*}

In this paper, we enable a robot to automatically understand the what, where, and how problems and propose a novel method for the robot to prepare a previously unseen real chair for a teddy bear agent to sit on, regardless of the initial pose of the chair.
In our method, as the robot has no prior knowledge of the object, it first reconstructs the object and reasons its sitting affordance in physics simulations.
A big challenge is that part of the object is always occluded when it is placed on a plane.
To fully reconstruct the object, the robot needs to rotate the object to a \textit{stable} pose which shows the occluded part, and fuse the images captured from different viewpoints accurately.
If the object is identified as a chair, it can be non-upright or inaccessible\footnote{A chair can be not sittable even though it is in its functional (upright) pose. For example, it can be blocked by obstacles (\textit{e.g.}, walls) and thus inaccessible for sitting. By accessible, we mean that the chair can be directly sitted on without any changes to its pose.}, and thus not sittable.
In this case, the robot needs to further rotate the object to a functional and accessible pose that is sittable for the bear if necessary.
In particular, we leverage the imagination method developed in \cite{wu2020chair, wu2021put} to guide the manipulation for object reconstruction and the manipulation for preparing the chair to a functional and accessible pose.
Fig. \ref{fig: pipeline} shows the pipeline of our method.

We state the difference between this work and our previous work \cite{wu2020chair, wu2021put} here.
In \cite{wu2020chair}, we proposed a method to identify the sitting affordance of an object (\textit{what}) and find the functional pose (\textit{where}) if it is classified as a chair.
No real robot experiments were performed.
In \cite{wu2021put}, we developed a method to find the sitting pose for an upright chair.
The object is known as a chair in an upright pose \textit{a priori} and no robot manipulation of the chair is included.
With human assistance, a NAO humanoid robot seats a teddy bear onto the chair.
In this paper, we address the understanding of sitting affordance from a real robot manipulation perspective.
We develop a robot manipulation system that enables the robot to manipulate the object automatically and intelligently, guided by the understanding of object stability and sitting affordance.
Instead of asking a human for assistance as in \cite{wu2021put}, we use a robot arm to automatically rotate an unseen chair from a randomly oriented pose to a functional pose. 
In the experiments, unlike \cite{wu2021put} which only has chairs in the test data, we include both chairs and non-chairs.
And the robot has no prior knowledge of the object category and whether it is placed in a functional (upright) pose or not.
The absence of these two kinds of prior knowledge brings great challenges, especially in the case when the test object is a flipped-over chair with its seat being occluded.

Our method successfully classifies 12 previously unseen objects with diverse shapes and appearances in 45 trials.
It achieves a 100\% success rate in preparing the chairs initially placed in a non-functional or inaccessible pose for sitting.
The humanoid robot seats the bear on the chair according to the imagined sitting poses, with a success rate of 96.7\%.
Comparing it with baselines on functional pose prediction, we empathize effectiveness of our imagination method.
The contributions of this paper mainly include:
\begin{itemize}
    \item a mathematical formulation of functional-equivalent and perceptible-equivalent poses.
    \item an automatic object reconstruction method guided by stability and perceptibility
    \item a real robot manipulation system for preparing an unseen chair for sitting.
\end{itemize}

\section{RELATED WORK}
\label{sec:related work}
\textbf{Object Affordance Reasoning.}
Affordance detection is attracting growing interest in both the fields of computer vision~\cite{deng20213d, myers2015affordance, zhu2014reasoning, ho1987representing, piyathilaka2019understanding} and robotics~\cite{do2018affordancenet, chu2019learning, wu2020can}. 
It helps robots with grasping~\cite{mandikal2021learning} and tool using~\cite{xu2021affordance, chu2019learning, myers2015affordance}.
Learning-based methods~\cite{xu2021affordance, do2018affordancenet} are popular for classifying objects and predicting affordance regions from images or point clouds.
But they are data intensive and the affordance information learned is insufficient for manipulation. 
As an alternative to learning, our method explores object affordances by simulating physical interactions.
This exploration of potential functionalities in simulation is what we refer to as \textit{robot imagination}.

\textbf{Physics Reasoning.}
Physical reasoning facilitates robot manipulation in a wide range of tasks including pouring~\cite{liang2015evaluating, wu2020can}, bottle opening~\cite{liu2019mirroring}, cutting~\cite{heiden2021disect}, and \textit{etc}.
Zhu \textit{et al.}~\cite{zhu2015understanding} learn physical concepts from an RGB-D video and pick the best tool for the task.
The problem of containability has been studied by physically simulating putting objects or particles into the object~\cite{liang2015evaluating, wu2020can}.
Liu \textit{et al.}\cite{liu2019mirroring} simulate robot actions and pick the one that causes the same effect as the human action they want.
Matl \textit{et al.}\cite{matl2020inferring} infer the material property of granular material using a physical simulator for planning pouring action and predicting material behavior.
This idea is similar to digital twins\cite{boschert2016digital} that is widely used in industry to predict system performance.
Our method goes beyond predicting outcomes and reasoning about interactions.
It explores object affordances by simulating objects interacting with humans and leverages the understanding on robot manipulation.

\textbf{Sitting Affordance Reasoning.}
Sitting is one of the most common postures of humans.
Hinkle and Olson\cite{hinkle2013predicting} drop spheres onto objects in simulation and classify the objects into chairs, tables, and containers according to the sphere's final configuration.
Grabner \textit{et al.}\cite{grabner2011makes} detect sitting affordance by fitting a humanoid mesh into the scene and evaluating the distance between the object and the humanoid.
Besides classification, we explore a deeper understanding of sitting and reason how an object is able to afford sitting and how to sit on it.

\section{PROBLEM FORMULATION}
\label{sec:problem_formulation}
The pose of a rigid body can be described as $g=(R, \mathbf{p}) \in SE(3)$,
where $R \in SO(3)$ is a rotation matrix that can be parameterized with zyx Euler angles $R = R_{ZYX}(\alpha, \beta, \gamma)=R_{Z}(\alpha)R_{Y}(\beta)R_{X}(\gamma)$.
$\alpha$, $\beta$, and $\gamma$ correspond to the yaw, pitch, and roll, respectively.
$R_{X}(\cdot)$, $R_{Y}(\cdot)$, and $R_{Z}(\cdot)$ are the rotation matrices representing rotations about the x-, y-, and z-axis of the world frame, respectively.
$\mathbf{p} = [x, y, z]^{T}\in\mathbb{R}^{3}$ represents the position.

We define an object as a chair if there exists any pose $g\in SE(3)$ that enables the object to afford the functionality of sitting.
When such a pose exists, it is a \textit{functional pose} of the object.
Given an unseen object in an arbitrary pose, our goal is threefold.
\textbf{1) Reconstruct a complete model of the object.}
When an object is placed on a plane, part of it is always occluded by the plane.
To scan the occluded part, we formulate the problem as rotating the chair into a \textit{stable} pose that shows the occluded part.
\textbf{2) Reason the sitting affordance from the reconstructed object model.}
This includes a) classifying whether an object is a chair (\textit{what}), b) finding the functional pose in which it can afford sitting (\textit{where}), and c) finding the sitting pose for the agent (\textit{how}).
\textbf{3) Prepare the chair for the bear.}
We use a teddy bear as a real agent to showcase the understanding of sitting.
Thus, the problem becomes manipulating the chair into a functional and accessible pose so that the teddy bear can be automatically seated by a humanoid robot.

\section{METHODS}
\label{sec:methods}
A rigid body has infinitely many possible poses in $SE(3)$.
It is intractable to search the whole $SE(3)$ space to find functional poses.
According to the IBD of chairs, a chair should be necessarily stable when an agent is sitting on it.
Therefore, we first find a discrete set of stable poses $G_{s} \subset SE(3)$ via \textit{stable pose imagination} (Sec. \ref{subsec: stable_pose_imagination}).
We then perform \textit{sitting imagination} (Sec. \ref{subsec: sitting_imagination}) on each stable pose $g_s \in G_{s}$ to check whether it can afford the sitting functionality.
If such a pose exists, the object is classified as a chair, and this pose is identified as a functional pose.

The teddy bear agent has almost rigid joints and the body configuration is very close to a sitting configuration.
Therefore, we simplify the problem of putting the bear onto a chair as finding the sitting pose $g_{sit} = (R_{sit}, \mathbf{p}_{sit})$ and leave the interaction of the agent as future work.
The IBD of chairs indicates that the agent's torso is more or less upright when sitting. 
We thus set the rotation of the agent as $R_{sit} = R_{Z}(\alpha_{sit})R_{0}$ in which the initial rotation $R_{0}$ puts the agent to an upright orientation.
We denote the direction the agent faces as the sitting direction.
The problem becomes finding one of the sitting position $\mathbf{p}_{sit}$ and the sitting direction indicated by the yaw angle $\alpha_{sit}$.

\subsection{Functionally and Perceptibly Equivalent Poses}
\label{subsec: functional_equivalent}
There are infinitely many stable poses of an object.
But we notice that many of them are \textit{equivalent} in terms of functionality and perceptibility.
For example, an upright chair can always afford the sitting functionality regardless of any planar motions (translation in the xy-plane and rotation about the z-axis).
When an object is placed on a plane, the upper side of the object can be directly perceived by the robot regardless of any planar motions.
Planar motions form a subgroup of $SE(3)$:
\begin{equation}
    H \doteq \{g \in SE(3) | g=(R, \mathbf{t}), R=R_{Z}(\alpha), \mathbf{t}=[x, y, 0]^{T}\} \cong SE(2)
    \label{eqn: transformation_subgroup}
\end{equation}
in which $\alpha\in[0, 2\pi)$, $x, y \in \mathbb{R}$.
And any $g\in SE(3)$ together with $H$ forms a coset\footnote{For brevity, we refer right cosets as cosets in this paper.} $Hg$\cite{chirikjian2011stochastic}:
\begin{equation}
    Hg = \{h \circ g: h \in H\} \subseteq SE(3)
\end{equation}
The equivalence of functionality and perceptibility is basically saying any two poses belonging to the same coset are equivalent in terms of functionality and perceptibility.
In other word, for any two poses $g_1=(R_1, \mathbf{p}_1)$ and $g_2=(R_2, \mathbf{p}_2)$, if there exists an $h \in H$ such that $g_2 = hg_{1}$, we say $g_{1}$ and $g_{2}$ are \textit{functionally equivalent} and \textit{perceptibly equivalent}.
We further define two rotations $R$ and $R'$ are functionally and perceptibly equivalent if there exists $\alpha \in [0, 2\pi)$ such that $R' = R_{Z}(\alpha)R$.
Otherwise, they are functionally and perceptibly unique.
The relative rotation matrix $R_{12}$ can be decomposed as $R_{12} = R_2R_1^{T} = R_z(\alpha_{12})R_{xy}(\phi_{12})$ where $R_{xy}(\cdot)$ is a rotation about an axis in the xy-plane and $\phi_{12}$ is the rotation angle.
In practice, we consider two poses to be functionally and perceptibly equivalent if:
\begin{equation}
   \phi_{12} < \Delta \phi_{es}, \quad |z_1 - z_2| < \Delta z_{es}
   \label{eqn: es_sim}
\end{equation}
$\Delta \phi_{es}$, and $\Delta z_{es}$ are two thresholds. 

We further define \textit{perceptible difference} $d_{pcp}$ to describe the perceptibility variation when the object is placed in two poses:
\begin{equation}
    d_{pcp}(g_1, g_2) = \phi_{12} = \arccos({\mathbf{n}_{z}^{T} R_{12}\mathbf{n}_{z}})
\end{equation}
where $\mathbf{n}_{z} = [0, 0, 1]^{T}$ is the z-axis unit vector.
And for any $hg_{1} \in Hg_{1}$ and $hg_{2} \in Hg_{2}$,
the perceptible difference are the same, \textit{i.e.}, $d_{pcp}(hg_{1}, hg_{2}) = d_{pcp}(g_{1}, g_{2})$.
Proof can be found in the \ref{sec:appendix} on our project page.

\subsection{Stable Pose Imagination}
\label{subsec: stable_pose_imagination}
In stable pose imagination, we simulate dropping an object in different initial poses onto a flat plane to find a set of stable poses $G_s$ as in \cite{wu2020chair}. 
For any $R \in SO(3)$, there exists a rotation $R' = R_{ZYX}(0, \beta, \gamma)$ that is functionally equivalent to $R$.
This is because $R$ can be decomposed as $R = R_{ZYX}(\alpha, \beta, \gamma) = R_{Z}(\alpha)R_{ZYX}(0, \beta, \gamma) = R_{Z}(\alpha)R'$.
That is, $R$ and $R'$ belong to the same coset, and thus show functional equivalence.
Therefore, for any stable pose $g_s=(R_s, \mathbf{p}_s)$, there exists a rotation $R_{ZYX}(0, \beta_{s}, \gamma_{s})$ which is functionally and perceptibly equivalent to $R_s$.
If the object is dropped from such a rotation in the simulation, it will very likely result in a pose that is functionally and perceptibly equivalent to $g_s$.
Thus, we enumerate the initial rotations of the object before dropping by varying the roll $\gamma$ and pitch $\beta$ while keeping $\alpha=0$.
For each dropping, we add the newly found stable pose to $G_s$ if it is functionally and perceptibly unique to all the poses in $G_s$.

We note that the stability of a pose $g$ is equivalent to all the poses in its coset $Hg$ because planar motions do not change stability.
That is, if $g$ is a stable pose, $hg$ is also a stable pose for any $h \in H$.
And finding a stable pose means finding a coset of stable poses.
If $g$ is a stable pose, we define $Hg$ as a stable pose coset.
We claim that given a sufficiently fine enumeration of $\beta$ and $\gamma$, the set of cosets formed by all the found stable poses $S = \{Hg_{s} | g_{s} \in G_{s}\}$ contains all the stable pose cosets of the object.
Proof can be found in \ref{sec:appendix}.

\subsection{Sitting Imagination}
\label{subsec: sitting_imagination}
\begin{figure}
    \centering
    \includegraphics[width=\columnwidth]{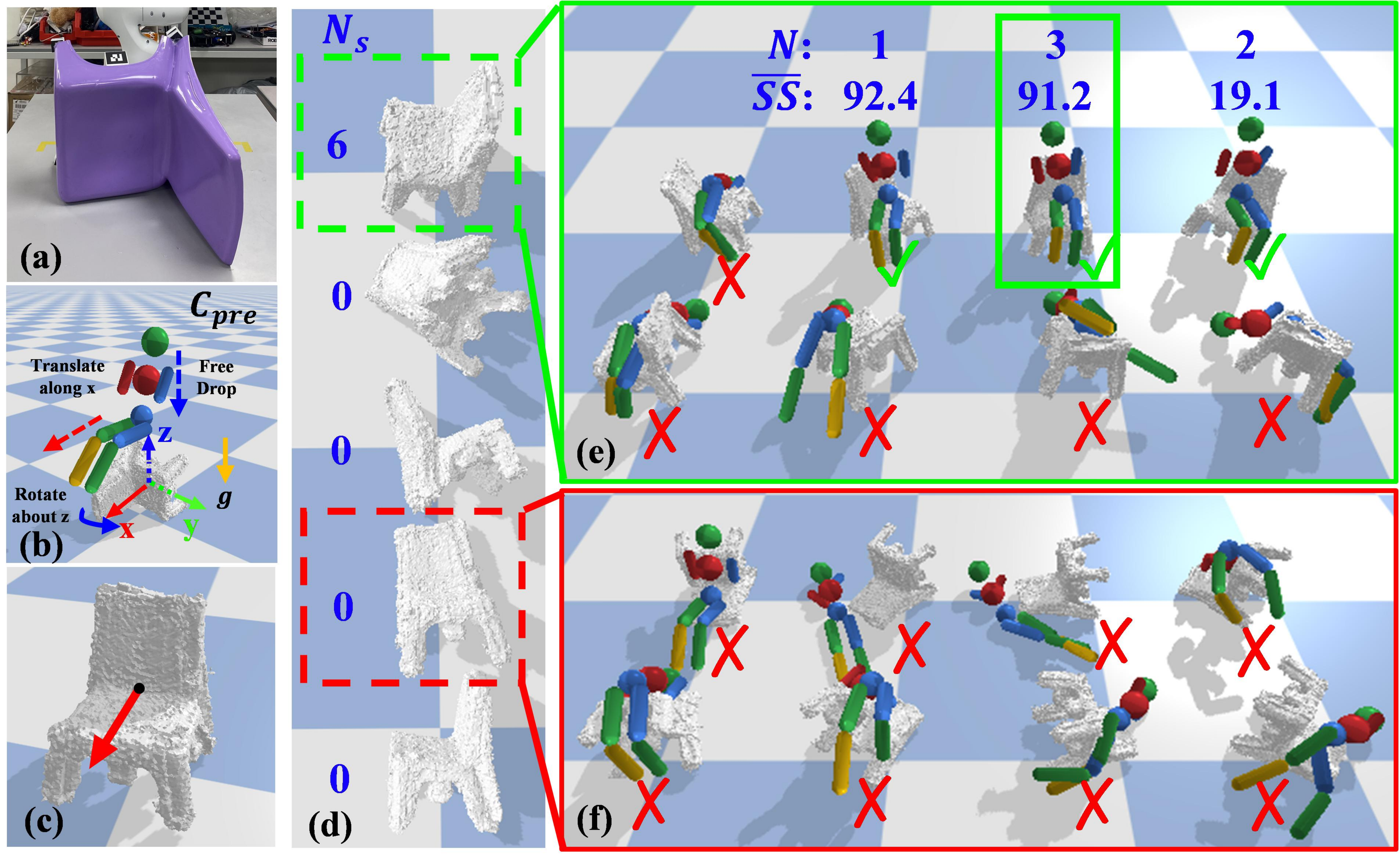}
    \caption{Sitting Imagination and Applications. (a) A test chair. (b) Sitting imagination setting. It shows the agent's initial configuration $C_{pre}$ before dropping. (c) The chair model is in a functional pose. The black dot shows the sitting position $\mathbf{p}_{sit}$; the red arrow shows the sitting direction $\alpha_{sit}$. (d) A set of stable poses $G_{s}$. The total number of correct sittings $N_s$ of each stable pose is shown on the left. (e) and (f) show the resultant configurations of sitting imagination of two stable poses. The correct sitting number $N$ and average sitting configuration score $\overline{SS}$ for each orientation are also shown. The stable pose in the green frame is recognized as a functional pose. In (e), the three sitting configurations with a check are correct sittings, with the framed one being the best sitting. The stable pose in the red frame is not a functional pose. In (f), there are no correct sittings.}
    \label{fig: imagination}
\end{figure}
To imagine the object sitting affordance, we simulate a passive agent sitting on the object with it in a stable pose.
We revisit the imagination method introduced in \cite{wu2020chair, wu2021put} briefly.
The human agent is modeled with an articulated human body with the forearms and feet trimmed off \cite{forssberg1994postural, kerr1997analysis}.
For each $g_s \in G_s$, we enumerate the orientations of the object by varying $\alpha$ in a discrete increment while fixing $\beta = \beta_s$, $\gamma = \gamma_s$.
The enumerated object poses are functionally equivalent to $g_s$.
We drop the agents onto the object with it in different enumerated orientations. 
Before each drop, we position the agent above the object and set it to a pre-sitting configuration $C_{pre}$ as shown in Fig. \ref{fig: imagination}(b).
The dropping position is enumerated along the positive x-axis of the object.
The position increment and the size of the agent $H_{agent}$ are scaled linearly to the size of the object.

For each drop, the agent resultant configuration $C_{res}$ is evaluated by the sitting affordance model (SAM) developed from \cite{wu2020chair, wu2021put} with five criteria. 
The first two criteria are similar to that in \cite{wu2020chair, wu2021put}. 
\textbf{1) Joint Angle.} 
The joint angle score $J$ is the weighted L1 distance between the joint angle vector of $C_{res}$ and a key configuration $C_{key}$.
Lower is better.
\textbf{2) Link Rotation.} 
Link rotation score $L$ is the weighted angular distance between the z-axis of all the links in $C_{res}$ and $C_{key}$.
Lower is better.
\textbf{3) Contact Points.} 
In IBD, the buttocks and back of the agent are supported by the object when it is sitting. 
Therefore, we consider the contact point number of the agent in SAM.
When sitting, the agent usually rests on the back with the chest and/or head in contact with the object.
In other cases, the back of a chair supports the body on the shoulders.
We count the upper body contact point $P_{upper} = P_{h} + P_{c}$ if $P_{h} + P_{c} > 0$; $P_{upper} = (P_{ls} + P_{rs}) / 2$ if otherwise.
$P_{h}$, $P_{c}$, and $P_{ls}$/$P_{rs}$ are the contact points of the head, chest, and left/right shoulder links, respectively.
The contact point number of the whole body is computed as $C = P_{upper} + P_{p} + P_{rt} + P_{lt}$ where $P_{p}$ and $P_{lt}$/$P_{rt}$ are the contact point of the pelvis and left/right thigh links, respectively.
\textbf{4) Symmetry.} Human bodies are generally symmetric when sitting. 
Therefore, we add the symmetry score $S$ to emphasize the body symmetry, \textit{i.e.}, $S = |j_{ls} - j_{rs}| + |j_{lt}-j_{rt}| + |j_{lk}-j_{rk}|$. 
$j_{ls}$/$j_{rs}$, $j_{lt}$/$j_{rt}$, and $j_{lk}$/$j_{rk}$ are the joint angles of the left/right shoulders, thighs and knees, respectively.
Lower is better.
\textbf{5) Sitting Height.} Sitting height is also an important factor in sitting.
The thigh height $H_t$ and pelvis height $H_p$ are the average height of all the contact points of the thighs and pelvis, respectively.
Lower is better.
We use a \textbf{Sitting Configuration Score} to measure the difference $SS = H_{t} / (J L H_{agent})$ between $C_{res}$ and $C_{key}$.
Higher is better.

A resultant agent configuration is considered as a \textit{correct sitting} if:
\begin{equation}
    J < J_{max},
    L < L_{max},
    C > C_{min}, 
    S < S_{max},
    SS > SS_{min}
    \label{eqn: score_condition}
\end{equation}
\begin{equation}
    P_{upper} > 0, P_{lt} > 0, P_{rt} > 0    
    \label{eqn: contact_condition_1}
\end{equation}
\begin{equation}
    H_{t} \in (H_{tmin}, H_{tmax}), H_{p} \in (H_{pmin}, H_{pmax}) 
    \label{eqn: sitting_height_condition_1}
\end{equation}
\begin{equation}
   |H_{p} - H_{t}| < \Delta H_{min}
    \label{eqn: sitting_height_condition_2}
\end{equation}
$J_{max}$, $L_{max}$, $S_{max}$, and $SS_{min}$ are thresholds corresponding to different scores.
We note that the object can support the agent body with fewer contact points when the agent is in a (nearly) symmetrical configuration.
Thus, we loosen the contact point threshold if the symmetry score is smaller than a more restricted threshold.
$H_{tmin}$, $H_{tmax}$, $H_{pmin}$, $H_{pmax}$, and $\Delta H_{min}$ are thresholds corresponding to the sitting height criteria.
Eqn. \ref{eqn: sitting_height_condition_2} gives a more strict restriction of the contact between the thigh and the object.
Considering a more comprehensive evaluation of contact points, sitting height, and symmetry, our method has a good performance when generalized to real data and the complicated application of chair preparation.
More details can be found in the \ref{sec:appendix}.

\subsection{Application of Imagination}
\label{subsec: application_of_imagination}
The imagination method introduced in Sec. \ref{subsec: stable_pose_imagination} and \ref{subsec: sitting_imagination} can be used for 1) object reconstruction, 2) chair v.s. non-chair classification, 3) functional pose prediction, and 4) finding the sitting pose of the agent.
Fig. \ref{fig: pipeline} shows the pipeline of our method.

Given an unseen object placed on a plane in a random pose $g$, the robot first scans the object and reconstructs an incomplete model as part of the object is occluded by the plane.
In order to reconstruct a complete object model, we need to find a stable pose that exposes the occluded part the most.
This means finding the stable pose coset which has the largest perceptible difference from the current pose.
Therefore, the robot performs stable pose imagination with this incomplete model to find a set of stable poses $G_s$.
The perceptible difference $d_{pcp}$ between each stable pose $g_{s} \in G_{s}$ and $g$ is calculated to find the stable pose $g_{s}^{pcp}$ with the largest $d_{pcp}$.
The poses in the stable pose coset $Hg_{s}^{pcp}$ offers the best perceptibility for the occluded part.
The robot manipulates the object to a pose $hg_{s}^{pcp} \in Hg_{s}^{pcp}$ that is perceptibly equivalent to this pose and scans the object again.
The complete object model is reconstructed by fusing this scan with the initial one.
More details can be found in \ref{subsec: robot_experiment_perception}.

Stable pose imagination is performed again on the complete object model to find the set of stable poses $G_s$.
The robot then performs sitting imagination on each $g_s \in G_{s}$ to check if any $g_{s}$ is a functional pose.
For every $g_{s}$, we count the number of correct sittings $N$ for each orientation $\alpha$.
We accumulate $N$ of all orientations as the correct sitting number $N_s$ for $g_s$. 
The stable pose $g_s$ that has the largest $N_s$ is selected as the candidate functional pose.
We classify an object as a sittable chair if the largest $N$ of the candidate functional pose satisfy:
\begin{equation}
   \max{N} > N_{min}
    \label{eqn: chair_classification_condition}
\end{equation}
$N_{min}$ is thresholds.
Otherwise, the object is classified as a non-chair.

If an object is recognized as a chair, the candidate functional pose is a functional pose denoted as $g_{f}$.
Each orientation $\alpha$ of $g_{f}$ generates an imagined sitting pose if the number of correct sittings $N>0$.
The sitting position $\mathbf{p}_{sit}$ and direction $\alpha_{sit}$ of this orientation is the weighted average of the agent base link position and yaw angle of the correct sittings.
The weight for each sitting is its corresponding $SS$.
And for each orientation, we compute the average sitting configuration score $\overline{SS}$ of all the correct sittings.
We rank the agent sitting poses by $N$.
If more than one sitting pose has the same $N$, we rank the one with the larger $\overline{SS}$ higher.
See Fig. \ref{fig: imagination} for an example.
More details on preparing the chair for sitting can be found in Sec. \ref{subsec: robot_experiment_pipeline}.

\section{EXPERIMENTS}
\label{sec:experiments}
\begin{figure}
    \centering
    \includegraphics[width=0.98\columnwidth]{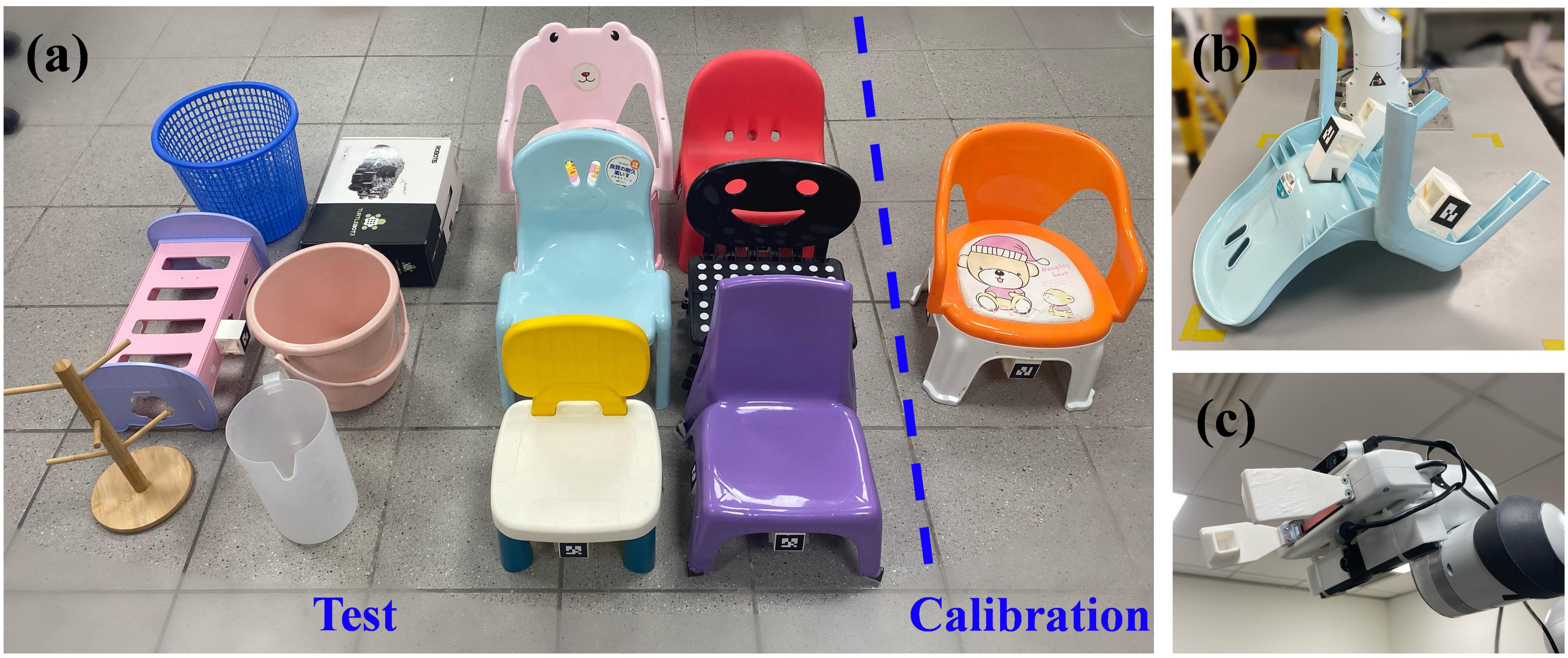}
    \caption{Experiment Details. (a) Robot Experiment Data. The right chair is used for calibrating the perception, imagination, planning, and control. The rest 12 objects are previously unseen and are used for testing our method and robot system. (b) A test chair with handles attached under the seat. (c) The gripper is mounted with two RGB-D cameras for scanning and tracking respectively.}
    \label{fig: experiment_details}
\end{figure}

Fig. \ref{fig: introduction}(a) shows the experiment setting.
The object is randomly placed on a table in front of the robot arm.
A Franka Emika robot arm is used to manipulate the object. 
Two RGB-D cameras are mounted on the end effector to scan the object and detect the grasping position, respectively (Fig. \ref{fig: experiment_details}(c)).
A NAO humanoid robot is used to carry the teddy bear and seat it on the chair.
An ArUro tag is placed on top of the NAO for tracking.

\subsection{Data}
\label{subec: robot_experiment_data}
The data of the experiment contains 7 real chairs and 6 non-chair objects that have different sizes, shapes, and appearances (Fig. \ref{fig: experiment_details}(a)).
The chairs are all designed for 0-3-year-old children.
We choose kiddie chairs because the size and weight of the chair are restricted by the workspace and payload of robots.
If the chair is too tall, the NAO is not able to reach the seat; if the chair is too heavy or large, the robot arm cannot manipulate it.

The imagination model introduced in Sec. \ref{sec:methods} is developed from that in \cite{wu2020chair}.
We calibrate our model using the calibration set of the synthetic dataset in \cite{wu2020chair}, which contains 30 synthetic chairs, and one real chair shown in Fig.\ref{fig: experiment_details}(a).
This real chair is also used for calibrating object reconstruction, motion planning, and control modules.
The rest chairs and all the non-chairs, which are unseen by the robot, are used as the test set.

\subsection{Robot Arm Manipulation}
We attach handles to all the objects to simplify the grasping problem in this paper.
An Alvar tag is attached to each handle for the robot to track the handle pose.
The robot has no prior knowledge of the number of handles and the handle poses.
It detects handles from the images captured for object reconstruction.
In particular, for each chair, we attach four handles underneath the seat.
We make sure the design and arrangement of the handles 1) do not change the object bounding box (OBB) of the chair, 2) do not affect the seat and back of the chair, 3) do not change the stable poses of the chair, and 4) guarantee that there is always at least one handle that is reachable by the robot arm.
After each manipulation, the grasped handle pose is tracked to provide an estimation of the object transformation in the manipulation.
IKFast \cite{diankov2010automated} is used to solve the inverse kinematics of the robot arm.
MoveIt \cite{chitta2012moveit} is used to plan the motion.
When putting the object down, we use a simple force controller, which terminates the downward motion of the arm when the vertical force exceeds a threshold.

\subsection{Object Reconstruction}
\label{subsec: robot_experiment_perception}
Part of the object is occluded when it is placed on the table.
In order to reconstruct a complete model, the robot manipulates the object to multiple perceptibly unique stable poses and scan it.
For each perceptibly unique stable pose, the robot arm moves to a set of pre-defined poses to capture depth images of the current scene.
We call this process a scan.
The point cloud of the scene is then reconstructed with TSDF fusion\cite{curless1996volumetric} and the object point cloud is segmented with plane segmentation.
The object transformation before and after manipulation is first estimated from the pose change of the handle and then refined by registering the object point clouds of both scans using iterative closest points (ICP)\cite{besl1992ICP}. 
The complete object model is reconstructed by integrating depth images captured from different scans.
Specifically, we transform the captured poses of all the depth images into the same frame with the estimated object transformations and use TSDF fusion to reconstruct the complete model.
To improve the quality of the complete model, we filter the depth image by removing the pixels corresponding to the table and the occluded part of the object as the depth values of the occluded part are usually noisy.

\subsection{Chair Preparation}
\label{subsec: robot_experiment_pipeline}
If the object is recognized as a chair, it can be non-upright or upright but inaccessible. 
In this case, the robot arm needs to manipulate the object to a functional \textit{and} accessible pose to prepare the chair for the NAO to seat the teddy bear.
In our setting, an upright chair is accessible if it is facing the NAO robot and inaccessible if otherwise (\textit{i.e.}, the chair faces the robot arm or the longer edge of the table).
We use the same method in \cite{wu2021put} to plan the $SE(2)$ trajectory to walk to the chair and the whole-body motion to seat the bear for the NAO.
In the experiments, for a functional pose, if the NAO fails to plan the walking trajectory or the whole-body motion, we consider it inaccessible.
To find a functional and accessible pose, the robot first tries to plan the motion to seat the bear on the chair with the highest-ranked imagined sitting pose.
If the planning is not successful, the robot tries the next-ranked imagined sitting pose.
If the planning is successful, the robot further plans the walking trajectories with the chair placed in different functional poses $hg_{f} \in Hg_{f}$.
The functional pose which results in successful trajectory planning is considered accessible.
The robot manipulates the chair to this pose and the NAO executes the motions to walk to the chair and seat the bear on the chair.

\begin{figure*}
    \centering
    \includegraphics[width=2\columnwidth]{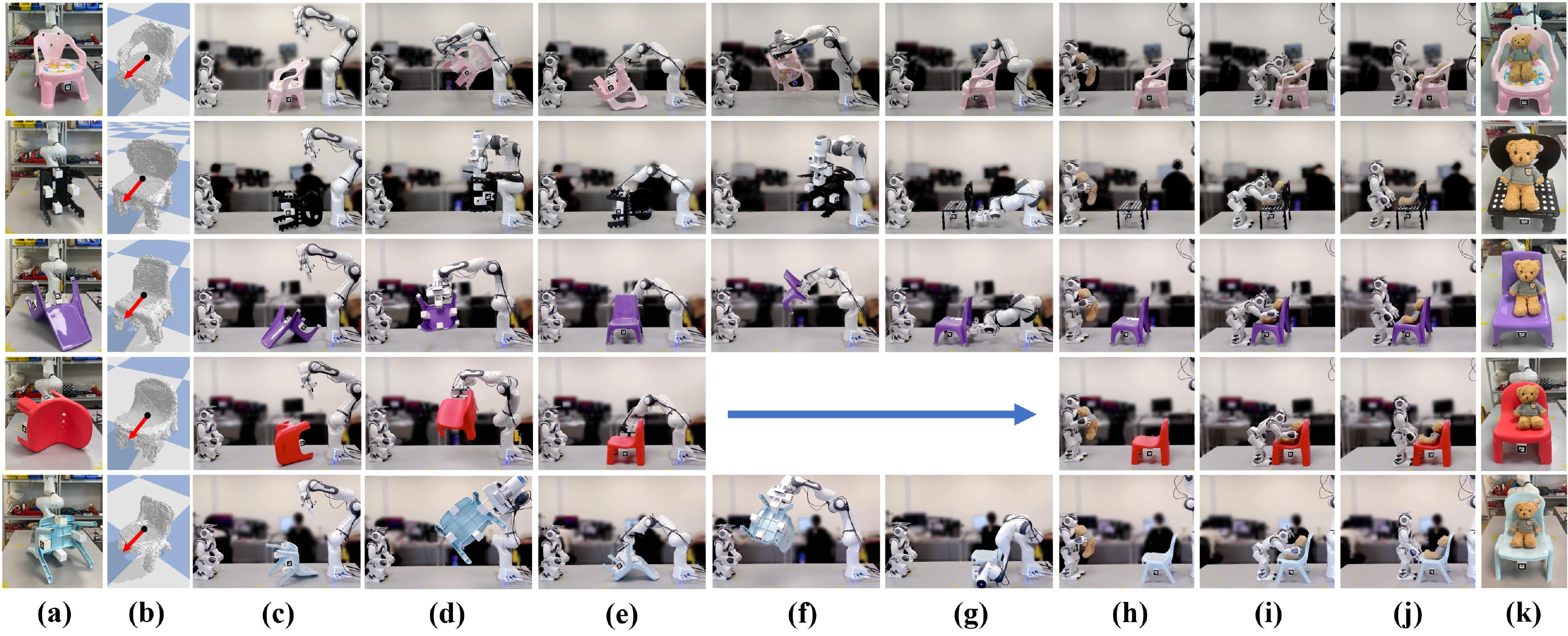}
    \caption{Real Robot Experiment Results. (a) Snapshot of the chair. (b) Imagined sitting pose. (c) Initial. (d)-(e) Rotating the chair to a perceptibly different pose to view the occluded part. (f)-(g) Preparing the chair to a functional and accessible pose.
    (h)-(j) Seating the bear on the chair. (k) Results.}
    \label{fig: robot_experiment}
\end{figure*}

\section{RESULTS}
\label{sec:results}
We test each object by placing it in some random poses that are functionally unique to each other.
In particular, the six unseen chairs are tested by placing them in five functionally unique poses.
In total, we perform 45 trials, including 30 trials of chairs and 15 trials of non-chairs, on the 12 objects in the test set.

\subsection{Annotation and Evaluation}
We recruit 5 volunteers to annotate the experiment result.
For each object in the test set, we ask the volunteers: "Do you think the object is a chair that is able to afford sitting?"
If the object is annotated as a chair, we further ask the volunteers to annotate the functional pose.
For each trial of the experiment, we first show the experiment video and image of the scene at the end of the experiment.
We then ask the volunteers: "Do you think the object is in an upright pose that is accessible for the NAO robot?"
If the answer to this question is positive, we further ask: "Do you think the robot has been successful in seating the bear on the chair?"
For each question, we consider the answer to the question positive if more than 3 out of 5 volunteers give a positive answer.

To evaluate the result of functional pose prediction, we compare the predicted pose with the annotated pose.
If they are functionally equivalent, \textit{i.e.} Eq \ref{eqn: es_sim} is satisfied with $\Delta v_{es} = 0.99, \Delta z_{es} = 0.01\mathrm{m}$, we consider the prediction correct.
If a chair object is classified as a non-chair, the prediction is considered incorrect.

\subsection{Functional Pose Prediction}
\subsubsection{Baseline}
We compare our method with five baseline methods on functional pose prediction of chairs.
\textbf{1) ICP Canonical.} Given the calibration chair as the canonical chair, we perform sitting imagination to find the functional pose, denoted as $(R_{cano}, p_{cano})$.
We use ICP to register the point cloud of unseen chairs to the point cloud of the canonical chair. 
The relative pose given by the registration is $R_{reg}$.
We apply the rotation $R_{cano}R_{reg}$ to the unseen chair, then drop the chair in simulation to find the functional pose.
\textbf{2) OBB Random.}
We randomly select one face of the object OBB and drop it with the selected face facing vertically downwards.
We propose this method because we observe that a chair often has one of the OBB faces in contact with the ground when it is in a functional pose.
\textbf{3) OBB Select.}
In this baseline method, we cast a ray from the center of each OBB face to the center of its opposite one.
We discard the faces with a collision-free ray.
The face of which the ray has the longest collision-free section before intersecting with the object is selected.
Besides the OBB-ground contact heuristic, we also note that the seat of a chair faces upwards and has no obstruction above when the chair is placed upright.
We use raycasting to check the existence of the seat and the free space above the seat.
For the first three baselines, the prior knowledge that the object is a chair is given.
The object is dropped using the same setting in stable pose imagination.
The resultant stable pose is the predicted functional pose.
\textbf{4) OBB Stable + Imagination}
We drop the object from the pose with each OBB face facing down to find a set of stable poses.
Sitting imagination is then performed on these stable poses to find the functional pose.
In this method, we replace stable pose imagination with OBB Stable to explore the effectiveness of stable pose imagination.
\textbf{5) Incomplete Reconstruction + Imagination}.
To study the effectiveness of complete object reconstruction with manipulation, we perform imagination on incomplete object models reconstructed from the initial scanning.
For these two baseline methods and our method, the prior knowledge that the object is a chair is not given.
If the object is classified as a non-chair, we consider the functional pose prediction unsuccessful.

\begin{table}[!htp]
\centering
\begin{tabular}{c c c}
\toprule
 Method & Prior & Success Rate (\%)\\
\midrule
ICP Canonical            & \Checkmark & 23.3 \\
OBB Random               & \Checkmark & 13.3 \\
OBB Select               & \Checkmark & 26.7 \\
\midrule
OBB Stable + Imagination & \XSolidBrush & 76.7 \\
Incomplete Object Model + Imagination & \XSolidBrush & 80.0 \\
\textbf{Imagination (Ours)}              & \XSolidBrush & \textbf{100.0} \\
\bottomrule
\end{tabular}
\caption{Results of Functional Pose Prediction}
\label{table: functional_pose_result}
\end{table}

\subsubsection{Results}
The results are shown in Tab. \ref{table: functional_pose_result}.
The performance of ICP canonical is relatively low.
A typical failure mode is registration failure.
One reason is that the shape and size variation of the chairs is large, making the registration fragile.
Another reason is the random orientation of chair models, bringing more challenges to registration.
The success rate of OBB Random is slightly lower than 1/6.
This verifies our heuristic that a chair has one of its OBB faces contacting the ground when it is in a functional pose.
But there are cases in which the heuristic is not correct, leading to a lower success rate.
OBB Select performs better by considering more heuristics.
However, the success rate is still not good.
OBB Stable + Imagination performs much better. 
Failure occurs when the set of stable poses found $G_{s}$ does not include a functional pose.
Compared with our method, it emphasizes the effectiveness of stable pose imagination.
The first five baseline methods are tested on the complete model of the chairs while Incomplete Object Model + Imagination is not.
It fails on 6 trials when the chair is overturned or lying on the side, resulting in poorly reconstructed seats.
Our full method achieves a 100\% success rate, outperforming all the comparing baseline methods.

\subsection{Real Robot Experiments}
We further evaluate our method on real robot experiments which includes four subtasks: 1) chair classification, 2) functional pose prediction, 3) chair preparation, and 4) seating the bear.
Qualitative results are shown in Fig. \ref{fig: robot_experiment}.
For chair classification, the robot achieves a 100\% success rate, correctly classifying the object in all 45 trials.
For chair preparation, the robot successfully prepares the chair to a functional and accessible pose in all the 30 trials with chairs.
In the 30 trials of chairs, the robot eventually succeeds on seating the bear on the chair in 29 trials, achieving a success rate of 96.7\%.
The failure case is a small chair without an armrest.
The highest-ranked imagined sitting pose is a pose in which the agent does not face the direction perpendicular to the back of the chair.
The control error and the restricted workspace of the NAO robot make it difficult to put the teddy bear precisely according to the imagined sitting pose.
Thus, the bear has its head somewhat supported by the back, which is not considered a success by 3 out of the 5 annotators.

\begin{figure}
\centering
\includegraphics[width=0.98\columnwidth]{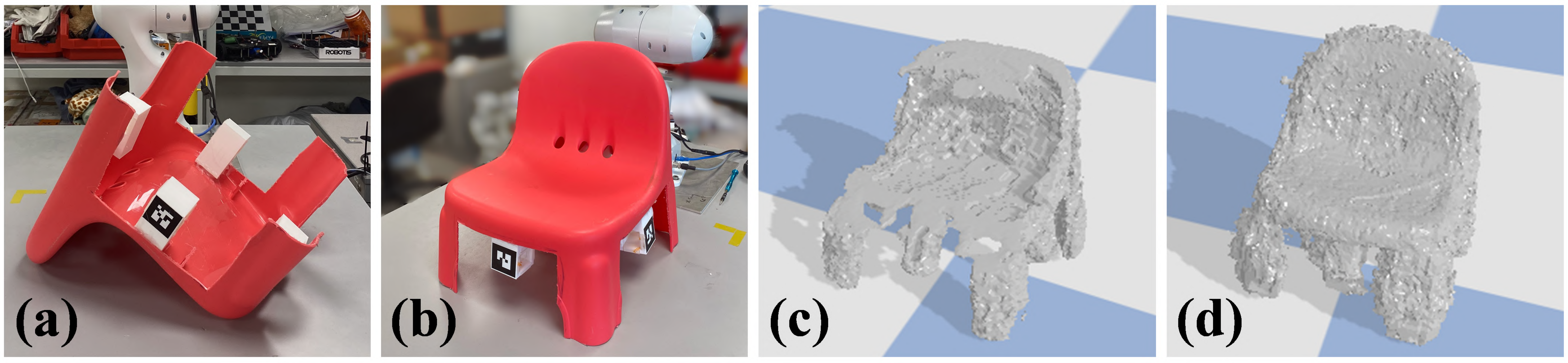}
\caption{Reconstruction Results. (a) Initial pose of a test chair. (b) The chair in an upright pose. (c) Incomplete model reconstructed from data captured when the chair is in the initial pose. (d) Complete model reconstructed by manipulation.}
\label{fig: reconstruction}
\end{figure}

\subsection{Object Reconstruction}
In Fig. \ref{fig: reconstruction}, we show an example of the incomplete object model and the complete model reconstructed using our method.
The complete model provides a more adequate representation of the seat and the back, allowing the robot to more accurately imagine the functionality and interact with the object physically.

\section{CONCLUSIONS \& FUTURE WORK}
\label{sec:conclusion}
In this letter, we proposed a novel method base on real2sim2real transfer in which robots imagine the sittability (affordance of chairs) of a previously unseen object and use this affordance-based reasoning to guide robot-object interaction.
We develop a robot manipulation system to actively perceive the object, prepare the chair for sitting and seat a teddy bear on the chair autonomously.
Results show that our method enables the robot to manipulate 6 novel chairs in 30 trials to a functional and accessible pose and seat the bear on them with a very high success rate.
Our method can be applied to various real-world tasks including tidying up the chairs, preparing chairs for a meeting, and can be possibly extended to help elderly people with sitting.
In the future, we plan to extend our imagination method to other objects that involve human whole-body interactions and explore more functionality-related physical properties through active interaction.






\bibliographystyle{IEEEtran}
\bibliography{references}
\newpage
\section*{SUPPLEMENTARY MATERIAL}
\label{sec:appendix}
\subsection{Proof for Rotation Enumeration}
Two rotations $R_1 \in SO(3)$ and $R_2 \in SO(3)$ are functionally and perceptibly equivalent if a $\alpha \in [0, 2\pi)$ exists such that $R_2 = R_z(\alpha)R_1$.
When combined with a random position $\mathbf{p} = [x, y, z]^T \in \mathbb{R}^3$, they form two poses, $g_1 = (R_1, \mathbf{p})$ and $g_2 = (R_2, \mathbf{p})$, that are functionally and perceptibly equivalent, \textit{i.e.}, $g_2 = hg_1$, where $h = (R_z(\alpha), [0, 0, 0]^T) \in H$.
We claim that for any pose $g' = (R', \mathbf{p}') \in SE(3)$, there exist a rotation $R_{ZYX}(0, \beta, \gamma)$ that enables the pose $g = (R, \mathbf{p}')$ functionally and perceptibly equivalent to $g'$, where $\beta \in [0, 2\pi)$, $\gamma \in [0, 2\pi)$.
We prove this by proving a functionally and perceptibly equivalent rotation $R_{ZYX}(0, \beta, \gamma)$ exists for any $R'$.
$R'$ can be decomposed using Euler angles as $R' = R_z(\alpha') R_y(\beta') R_x(\gamma')$.
There always exists a $R = R_y(\beta') R_x(\gamma')$ such that $R' = R_z(\alpha')R$, \textit{i.e.}, $R$ and $R'$ are functionally and perceptibly equivalent.
By enumerating $\beta$, $\gamma$ and $\mathbf{p}$, we are able to cover all cosets in $SE(3)$, \textit{i.e.}, we are able to enumerate a pose that is functionally and perceptibly equivalent to any $g' \in SE(3)$.

For any stable pose $g_s = (R_s, \mathbf{p}_s)$, if we position the object in a position $\mathbf{p}_i$ above the plane, and an orientation $R_s$, by dropping the object in such a position, it will result in $g_s$.
In our enumeration, we vary $\beta$ and $\gamma$ from $[0, 2\pi)$ in discrete increments $\Delta\beta$ and $\Delta\gamma$.
By keeping $\mathbf{p} = \mathbf{p}_i$, our enumeration is able to find a pose that is functionally and perceptibly equivalent to (or close enough to a pose that is functionally and perceptibly equivalent to) $(R_s, \mathbf{p}_i)$.
By dropping the object in such a pose, the object will be very likely to result in a pose that is functionally and perceptibly equivalent to $g_s$.
That is, by dropping the object from all poses enumerated by our enumeration, we are able to find a set of all functionally and perceptibly unique poses $G_s$.
The set of cosets formed by all the found stable poses $S = \{Hg_{s} | g_{s} \in G_{s}\}$ contains all the stable pose cosets of the object.

\subsection{Proof for Perceptible Difference}
Given two random poses $g_1 = (R_1, \mathbf{p}_1) \in SE(3)$ and $g_2 = (R_2, \mathbf{p}_2) \in SE(3)$, we claim that for any $hg_1 \in Hg_1$ and $hg_2 \in Hg_2$, the perceptible difference are the same, \textit{i.e.}, $d_{pcp}(hg_1, hg_2) = d_{pcp}(g_1, g_2)$.

Proof:
\begin{equation*}
    \begin{aligned}
    d_{pcp}(hg_1,hg_2)&=\arccos[n_z^T(R_z(\alpha)R_2(R_z(\alpha)R_1)^T)n_z]\\
    &= \arccos[n_z^TR_z(\alpha)(R_2R_1^T)R_z(\alpha)^T n_z]\\
    &= \arccos [n_z^T(R_2R_1^T) n_z]\\
    &= d_{pcp}(g_1,g_2)
    \end{aligned}
\end{equation*}
where $R_Z n_Z = n_Z$ is used.

\subsection{Experiment Details}
\subsubsection{Imagination Details}
Following the settings of our previous papers, in stable pose imagination, the initial orientation of the object is enumerated by varying $\beta$ and $\gamma$ in 15 discrete values within $[0, 2\pi)$, \textit{i.e.}, $\Delta \beta = \Delta \gamma = 2 \pi / 15$.
In each sitting imagination, the agent is dropped onto the chair in 8 different orientations, \textit{i.e.}, $\Delta \alpha = \pi / 4$.

The SAM classifies and evaluates the configuration of an agent based on five criteria, which range from coarse to fine.  
The joint score and link score provide an overall comparison with the key configuration, while the other three focus on different specific aspects. 
The evaluation of contact points ensures proper support for the agent's body. 
The symmetry score takes into account the intuition that natural sitting configurations for humans are typically symmetrical. 
The height score focuses on the seat's height and flatness, ensuring that it is comfortable for human use. 
The thresholds are: $J_{max} = 2.0$, $L_{max} = 0.5$, $C_{min} = 4$, $S_{max} = 2.0$, $SS_{min} = $, $H_{tmin} = H_{pmin} = 0.15 H_{agent}$, $H_{tmax} = H_{pmax} = 0.15 H_{agent} + 1.0$, $\Delta H_{min} = 0.15 H_{agent} $.

In functional pose prediction, the thresholds $N_{min} = 3$.

\subsubsection{Results Details}
\begin{figure}
    \centering
    \includegraphics[width=0.9\columnwidth]{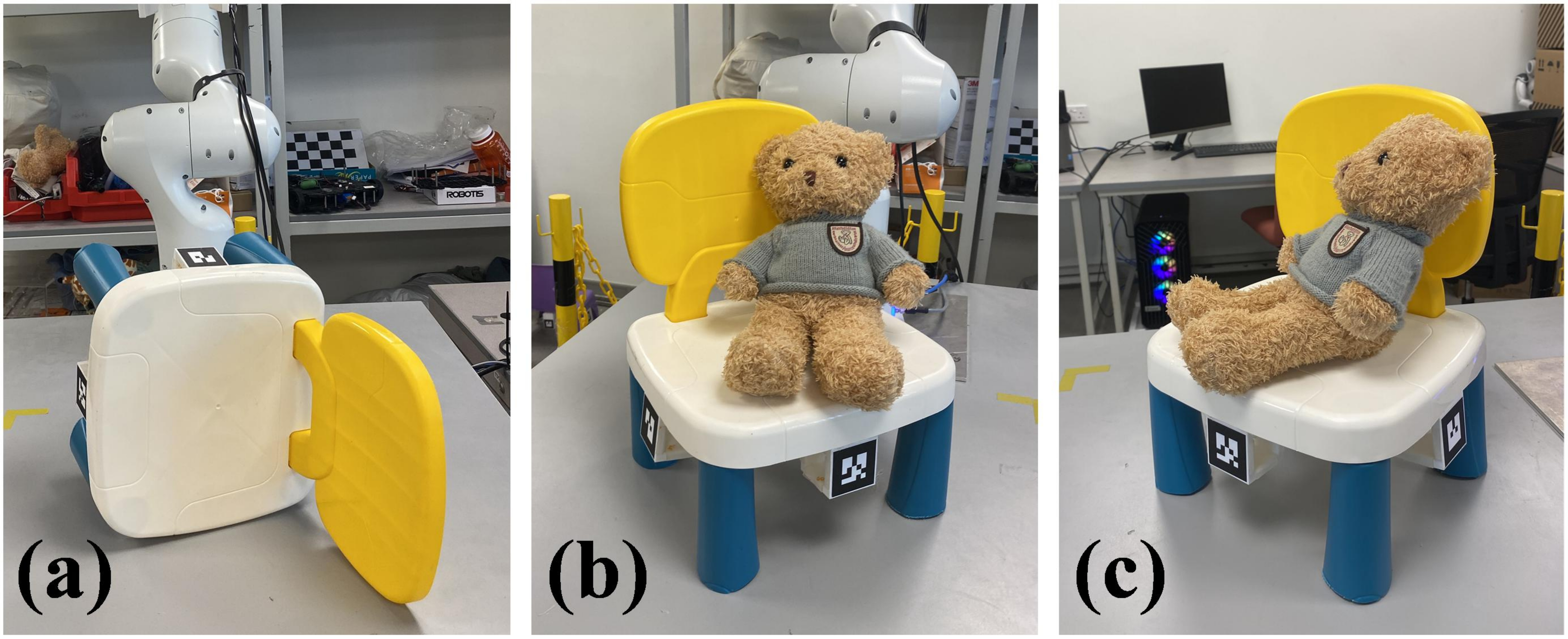}
    \caption{Failure Case. (a) Initial pose of a test chair. (b) and (c) shows the result of chair preparation and bear seating. }
    \label{fig: failure_case}
\end{figure}
In the real experiments, our method achieved a 100 \% success rate in chair classification and functional pose prediction of the chairs. 
In the 30 trials of chairs, the robot successfully seats the bear on the chair in 29 trials.
The failure case is shown in Fig. \ref{fig: failure_case} where the test chair doesn't have an armrest. 
Among the imagined sitting poses, the highest-ranked sitting direction is not (nearly) perpendicular to the back of the chair.
The sitting position is difficult for NAO robot to reach from such a direction.
The control error and the restricted workspace of NAO also make it difficult to precisely put the teddy bear into the imagined pose.
Thus, the bear only has its head supported by the chair back from the side.
Three out of five annotators do not regard it as a successful sitting.
Because the back and/or the head of the bear are not properly supported, the bear is likely to fall down from the chair.

\subsection{Glossary}
\subsubsection{Calibration}
In this paper, \textit{calibration} refers to the process of determining and adjusting the parameters in a module.
Calibration data refers to the data used in the process of calibration.
In imagination, the goal of calibration is to ensure that the imagination provides a reliable representation of actions and an accurate evaluation of the resultant configuration.
For example, we use the calibration data to tune the dynamic parameters of the simulation environment to decrease discrepancies between the simulated actions and real-world observations.
In real robot system, the goal of calibration is to improve the accuracy of reconstruction, motion planning, and control.
For example, in reconstruction, we adjust the value of thresholds in plane segmentation to make sure the table is correctly removed.

\subsubsection{Perceptibility}
\textit{Perceptibility} refers to the capability of being perceived by the senses.
More specifically, it refers to the degree to which an object can be perceived by the robot in a certain configuration.
For example, when a chair is placed upright on a table, the bottoms of the legs are occluded, and the rest part can be captured, \textit{i.e.}, perceptible.
Whereas when it is overturned, part of the seat and back is occluded, and the rest sections are perceptible.
The perceptibility of the chair in such two different poses is different.
We use the \textit{perceptible difference} to describe this variation in perceptibility.

\end{document}